\documentclass[runningheads]{llncs}
\usepackage{graphicx}
\usepackage{amssymb}
\usepackage{url}
\usepackage{amssymb}
\usepackage{amsmath, xparse}

\usepackage[caption=false]{subfig} %https://tex.stackexchange.com/questions/499870/lecture-notes-in-computer-science-lncs-side-by-side-figures
\usepackage[export]{adjustbox}
\begin{document}
\title{An Open-Source and Reproducible Implementation of LSTM and GRU Networks for Time Series Forecasting}

\titlerunning{LSTM and GRU Networks for Time Series Forecasting}

%ORCID\orcidID{0000-0001-5392-9540} 
%\author{Gissel Velarde\thanks{Contact author: Gissel Velarde\orcidID{0000-0001-5392-9540}} \and
\author{Gissel Velarde\orcidID{0000-0001-5392-9540} \and
Pedro Bra\~{n}ez \and
Alejandro Bueno \and \\
Rodrigo Heredia \and
Mateo Lopez-Ledezma}
\authorrunning{G. Velarde et al.}

\institute{Independent, Bolivia  \\
gv@urubo.org, pedrobran8@gmail.com, alebuenoaz@gmail.com, rodrigoh1205@gmail.com, lopezmateo97@yahoo.com } 

\maketitle              % typeset the header of the contribution
\begin{abstract}
This paper introduces an open-source and reproducible implementation of  Long Short-Term Memory (LSTM) and Gated Recurrent Unit (GRU) Networks for time series forecasting.  We evaluated LSTM and GRU networks because of their performance reported in related work. We describe our method and its results on two datasets. The first dataset is the S\&P BSE BANKEX, composed of stock time series (closing prices) of ten financial institutions. The second dataset, called Activities, comprises ten synthetic time series resembling weekly activities with five days of high activity and two days of low activity. We report Root Mean Squared Error (RMSE) between actual and predicted values, as well as Directional Accuracy (DA). We show that a single time series from a dataset can be used to adequately train the networks if the sequences in the dataset contain patterns that repeat, even with certain variation, and are properly processed. For 1-step ahead and 20-step ahead forecasts, LSTM and GRU networks significantly outperform a baseline on the Activities dataset. The baseline simply repeats the last available value. On the stock market dataset, the networks perform just like the baseline, possibly due to the nature of these series. We release the datasets used as well as the implementation with all experiments performed to enable future comparisons and to make our research reproducible. 

\keywords{Forecasting  \and Time Series \and Open-source \and Reproducibility.}

\end{abstract}
\section{Introduction}
Artificial Neural Networks (ANNs) and particularly Recurrent Neural Networks (RNNs) gained attention in time series forecasting due to their capacity to model dependencies over time \cite{rumelhart1986learning}. With our proposed method,\footnote{The code and datasets are available at: \\
\url{https://github.com/Alebuenoaz/LSTM-and-GRU-Time-Series-Forecasting} \cite{velarde2022a}.} we show that RNNs can be successfully trained with a single time series to deliver forecasts for unseen time series in a dataset containing patterns that repeat, even with certain variation. Therefore, once a network is properly trained, it can be used to forecast other series in the dataset if adequately prepared. 

LSTM \cite{hochreiter1997long} and GRU \cite{cho2014properties} are two related deep learning architectures from the RNN family. LSTM consists of a memory cell that regulates its flow of information thanks to its non-linear gating units, known as the input, forget and output gates, and activation functions \cite{greff2016lstm}. GRU architecture consists of reset and update gates, and activation functions. Both architectures are known to perform equally well on sequence modeling problems, yet GRU was found to train faster than LSTM on music and speech applications \cite{chung2014empirical}.

Empirical studies on financial time series data reported that LSTM outperformed Autoregressive Integrated Moving Average (ARIMA) \cite{siami2018comparison}. ARIMA \cite{box_jenkins1970} is a traditional forecasting method that integrates autoregression with moving average processes. In \cite{siami2018comparison}, LSTM and ARIMA were evaluated on RMSE between actual and predicted values on financial data.  The authors suggested that the superiority of LSTM over ARIMA was thanks to gradient descent optimization \cite{siami2018comparison}. A systematic study compared different ANNs architectures for stock market forecasting \cite{balaji2018applicability}. More specifically, the authors evaluated architectures of the types LSTM, GRU, Convolutional Neural Networks (CNN), and Extreme Learning Machines (ELM). In their experiments, two-layered LSTM and two-layered GRU networks delivered low RMSE. 

In this study, we evaluate LSTM and GRU architectures because of their performance reported in related work for time series forecasting \cite{siami2018comparison,balaji2018applicability}. Our method is described in section  \ref{s:method}. In sections \ref{s:RNN}, \ref{s:LSTM}, and \ref{s:GRU}, we review principles of Recurrent Neural Networks (RNN) of the type LSTM and GRU.   In section \ref{s:datap}, we explain our data preparation, followed by the networks' architecture, training (section \ref{s:at}), and evaluation (section \ref{s:eva}). The evaluation is done on two datasets. In section \ref{s:BANKEX}, we describe the S\&P BSE-BANKEX or simply BANKEX dataset, which was originally described in \cite{balaji2018applicability} and consists of stock time series (closing prices). In section \ref{s:activities}, we describe the Activities dataset, a dataset composed of synthetic time series resembling weekly activities with five days of high activity and two days of low activity. The experiments are presented in section \ref{s:experiments}.  Finally, we state our conclusions in section \ref{s:conclusions}, and present possible directions for future work. We release the datasets used as well as the implementation with all experiments performed to enable future comparisons and make our research reproducible.

\section{METHOD} \label{s:method}
The general overview of the method is described as follows. The method inputs time series of values over time and outputs predictions. Every time series in the dataset is normalized. Then, the number of test samples is defined to create the training and testing sets. One time series from the train set is selected and prepared to train a LSTM and a GRU, independently. Once the networks are trained, the test set is used to evaluate RMSE and DA between actual and predicted values for each network. The series are transformed back to unnormalized values for visual inspection. We will describe every step in detail. Next, in sections \ref{s:RNN}, \ref{s:LSTM}, and \ref{s:GRU}, we review principles of  RNNs of the type LSTM and GRU, following the presentation as in  \cite{chung2014empirical}.

\begin{figure*}
   \includegraphics[width=1\linewidth]{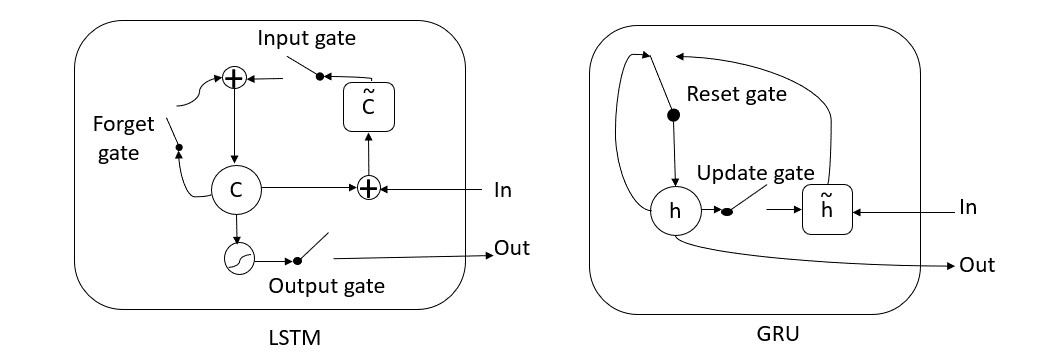}
   \caption{LSTM (left) and GRU (right). $c$ represents the memory cell and $\tilde{c}$ the new memory cell of the LSTM. $h$ represents the activation and $\tilde{h}$ the new activation of the GRU. Based on \cite{chung2014empirical}.}
   \label{figLSTM-GRU} 
\end{figure*}

\subsection{Recurrent Neural Networks} \label{s:RNN}
ANNs are trained to approximate a function and learn the networks' parameters that best approximate that function. RNNs are a special type of ANNs developed to handle sequences. A RNN updates its recurrent hidden state $h_t$ for a sequence $x= (x_1, x_2, ..., x_T)$ by:
\begin{equation} \label{eq1}
h_t=\left\{\begin{matrix}
0, & t=0 \\        
\phi (h_{t-1},x_t), & otherwise,
\end{matrix}\right.
\end{equation}

where $\phi$ is a nonlinear function. The output of a RNN maybe of variable length $y=(y_1, y_2, ..., y_T)$.

The update of $h_t$ is computed by:
\begin{equation}
    h_t = g(Wx_t+Uh_{t-1}),
\end{equation}
where $W$ and $U$ are weights' matrices and $g$ is a smooth and bounded activation function such as a logistic sigmoid, or simply called sigmoid function $f(x)=\sigma=\frac{1}{1+e^{-x}}$, or a hyperbolic tangent function $f(x)=tanh(x)=\frac{e^{x}-e^{-x}}{e^{x}+e^{-x}}$.
 
Given a state $h_t$, a RNN outputs a probability distribution for the next element in a sequence. The sequence probability is represented as:
\begin{equation}
    p(x_1, x_2, ..., x_T)= p(x_1)p(x_2\mid  x_1)...p(x_T\mid x_1, x_2,...,x_{T-1}).
\end{equation}
 The last element is a so called end-of-sequence value. The conditional probability distribution is given by:
 \begin{equation}
     p(x_t\mid x_1, x_2,...x_{t-1})=g(h_t),
 \end{equation}
where $h_t$ is the recurrent hidden state of the RNN as in expression (\ref{eq1}). Updating the network's weights involves several matrix computations, such that back-propagating errors lead to vanishing or exploding weights, making training unfeasible. LSTM was proposed in 1997 to solve this problem by enforcing constant error flow thanks to gating units \cite{hochreiter1997long}. GRU is a closely related network proposed in 2014 \cite{cho2014properties}. Next, we review LSTM and GRU networks. See Figure \ref{figLSTM-GRU} for illustration.

\subsection{Long Short-Term Memory}\label{s:LSTM}
The LSTM unit decides whether to keep content memory thanks to its gates. If a sequence feature is detected to be relevant, the LSTM unit keeps track of it over time, modeling dependencies over long-distance \cite{chung2014empirical}. 

In expressions  (\ref{eq:6}), (\ref{eq:9}), (\ref{eq:10}), and (\ref{eq:8}), $W$ and $U$ represent weights matrices and $V$ represents a diagonal matrix. $W$, $U$ and $V$ need to be learned by the algorithm during training. The subscripts $i$, $o$, and $f$ correspond to input, output, and forget gates, respectively. For every $j$-th LSTM unit, there is a memory cell $c_{t}^{j}$ at time $t$, which activation $h_t^{j}$ is computed as:
\begin{equation}
  h_{t}^{j} = o_{t}^{j} tanh(c_t^{j})
\end{equation}

where $o_t^{j}$ is the output gate responsible for modulating the amount of memory in the cell. The forget gate $f_t^{j}$ modulates the amount of memory content to be forgotten and the input gate $i_{t}^{j}$ modulates the amount of new memory to be added to the memory cell, such that:

\begin{equation}  \label{eq:6}
    o_{t}^{j} = \sigma(W_ox_t + U_o h_{t-1} + V_o c_t)^{j},
\end{equation}

\begin{equation}  \label{eq:9}
f_{t}^{j} = \sigma(W_fx_t + U_f h_{t-1} + V_f c_{t-1})^{j}, 
\end{equation}

\begin{equation}  \label{eq:10}
i_{t}^{j} = \sigma(W_ix_t + U_i h_{t-1} + V_i c_{t-1})^{j}.
\end{equation}
 
where $\sigma$ is a sigmoid function. The memory cell $c_{t}^{j}$ partially forgets and adds new memory content $\tilde{c}_t^{j}$ by:
\begin{equation}
c_{t}^{j}=f_t^{j}c_{t-1}^{j}+i_t^{j} \tilde{c}_t^{j},
\end{equation}

where:
\begin{equation}  \label{eq:8}
\tilde{c}_{t}^{j}=tanh(W_cx_t + U_c h_{t-1})^{j}.
\end{equation}

\subsection{Gated Recurrent Unit}\label{s:GRU}
The main difference between LSTM and GRU is that GRU does not have a separate memory cell, such that the activation $h_t^{j}$ is obtained by the following expression:
\begin{equation}
h_t^{j} = (1-z_t^{j}) h_{t-1}^{j}+ z_t^{j} \tilde{h}_{t}^{j}.
\end{equation}

 The update gate $z_t^{j}$ decides the amount of update content given by the previous $h_{t-1}^{j}$ and candidate activation $\tilde{h}_{t}^{j}$. In expressions (\ref{eq:12}), (\ref{eq:13}), and (\ref{eq:14}), $W$ and $U$ represent weights matrices that need to be learned during training. Moreover, the subscripts $z$ and $r$ correspond to update and reset gates, respectively. The update gate $z_t^{j}$ and reset gate $r_t^{j}$ are obtained by the following expressions:
\begin{equation}  \label{eq:12}
z_t^{j} = \sigma(W_z x_{t}+ U_z h_{t-1})^{j}, 
\end{equation}

\begin{equation} \label{eq:13}
r_t^{j} = \sigma(W_r x_{t}+ U_r h_{t-1})^{j},
\end{equation}

where $\sigma$ is a sigmoid function. The candidate activation $\tilde{h}_t^{j}$ is obtained by:  
\begin{equation} \label{eq:14}
   \tilde{h}_t^{j} = tanh(W x_{t}+ r_t \odot (U h_{t-1}))^{j}. 
\end{equation}
where $\odot$ denotes element-wise multiplication. 

\subsection{Data Preparation}\label{s:datap}
Every time series or sequence in the dataset is normalized as follows. Let be $v$ a sequence $v= (v_1, v_2, ..., v_Q)$ of $Q$ samples that can be normalized between 0 and 1: 

\begin{equation} \label{norma}
x=v'=\frac{v-v_{min}}{v_{max}-v_{min}}.
\end{equation}

We define the number of samples in the test set as $test_s$. The number of samples $N$ for training is obtained by $N=Q-test_s-w$. Then, a sequence $x$ is selected arbitrarily and prepared to train each network as follows. We define a window of size $w$ and a number of steps ahead $f$, where $f<w< N < Q$, such that:
\begin{equation*}
X = \begin{bmatrix}
x_1 & x_2 & ... & x_w\\ 
x_2 & x_3 & ... & x_{w+1}\\ 
x_3 & x_4 & ... & x_{w+2}\\
... & ... & ... & ...\\ 
x_{Q-(w-1+f)} & x_{Q-(w-2+f)} & ... & x_{Q-f}\\ 
\end{bmatrix},
\end{equation*}
$X$ becomes a $Q-(w-1+f)$ by $w$ matrix, and:
\begin{equation}
Y = \begin{bmatrix}
x_{w+1}&x_{w+2}& ... & x_{w+f}\\ 
x_{w+2}&x_{w+3}& ... & x_{w+2+f}\\ 
x_{w+3}&x_{w+4}& ... & x_{w+3+f}\\
... & ... & ... & ...\\ 
x_{Q-(f-1)} & x_{Q-(f-2)} & ... & x_{Q}\\ 
\end{bmatrix}
\end{equation}
becomes a $Q-(w-1+f)$ by $f$ matrix containing the targets. The first $N$ rows of $X$ and $Y$ are used for training. The remaining $Q-N$ elements are used for testing. The settings for our experiments are described in section \ref{s:experiments}, after we introduce the characteristics of the dataset used. 

\subsection{Networks' architecture and training}\label{s:at}
We tested two RNNs. One with LSTM memory cells and one with GRU memory cells. In both cases, we use the following architecture and training:
\begin{itemize}
\item A layer with 128 units, 
\item A dense layer with size equal to the number of steps ahead for prediction, 
\end{itemize}

with recurrent sigmoid activations and $tanh$ activation functions as explained in sections \ref{s:LSTM} and \ref{s:GRU}. 
The networks are trained for 200 epochs with Adam optimizer \cite{kingma2014adam}. 
The number of epochs and architecture were set empirically. We minimize Mean Squared Error (MSE) loss between the targets and the predicted values, see expression (\ref{e:1}). The networks are trained using a single time series prepared as described in section \ref{s:datap}. The data partition is explained in section \ref{s:partition}.

\subsection{Evaluation}\label{s:eva}
We use Mean Squared Error (MSE) to train the networks:
\begin{equation}  \label{e:1}
MSE = n^{-1}\sum_{t=1}^{n}(x_t-y_t)^2,
\end{equation}

where $n$ is the number of samples, $x_t$ and $y_t$ are actual and predicted values at time $t$. Besides, we use Root Mean Squared Error (RMSE) for evaluation between algorithms: 
\begin{equation}
RMSE = \sqrt{MSE}.
\end{equation}

Both metrics, MSE and RMSE are used to measure the difference between actual and predicted values, and therefore, smaller results are preferred  \cite{wang2012stock}. We also use Directional Accuracy (DA):

\begin{equation}
DA = \frac{100}{n}\sum_{t=1}^{n}d_t,
\end{equation}
where:
\begin{equation*}
d_t=\left\{\begin{matrix}
1 & (x_{t}-x_{t-1})(y_{t}-y_{t-1})\geq 0\\ 
0 & otherwise.
\end{matrix}\right.
\end{equation*}

such that $x_t$ and $y_t$ are the actual and predicted values at time $t$, respectively, and $n$ is the sample size. DA is used to measure the capacity of a model to predict direction as well as prediction accuracy. Thus, higher values of DA are preferred \cite{wang2012stock}. 

\section{Experiments}\label{s:experiments}
In this section, we report experiments performed with both datasets.

\begin{figure}[htp]
\centering
\subfloat[Without normalization\label{fig:subim1}]{%
  \includegraphics[width=0.94\textwidth]{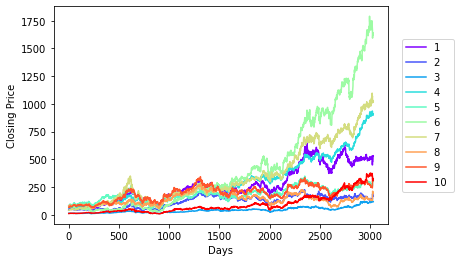}%
}\vfil
\subfloat[With normalization\label{fig:subim2}]{%
  \includegraphics[width=0.9\textwidth]{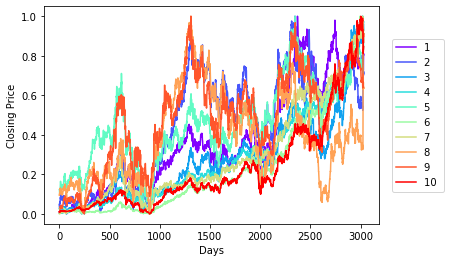}%
}

\caption{(a) Time series in the BANKEX dataset without normalization. Closing Price in Indian Rupee (INR). Daily samples retrieved between July 12, 2005 and November 3, 2017 using Yahoo! Finance's API \cite{YahooF_API}. All time series with $3\,032$ samples. (b) Same time series as in (a). Closing Price normalized between 0 and 1. The numbers from 1 to 10 correspond to the numbers (first column) for each series in Table \ref{tab1}.}
\label{fig:norm}

\end{figure}

\subsection{The S\&P BSE BANKEX Dataset} \label{s:BANKEX}
This dataset was originally described in \cite{balaji2018applicability}, however, our query retrieved a different number of samples as in \cite{balaji2018applicability}. We assume it must have changed since it was originally retrieved. We collected the time series on January 20, 2022, using Yahoo! Finance's API \cite{YahooF_API} for the time frame between July 12, 2005, and November 3, 2017, see Table \ref{tab1}.  Most time series had $3\,035$ samples, and some time series had $3\,032$ samples. Therefore, we stored each time series's last $3\,032$ samples. Figure \ref{fig:norm} presents the time series of BANKEX without and with normalization.   
\begin{table}[]
\centering
\caption{Entities in the S\&P BSE-BANKEX Dataset.}\label{tab1}
\begin{tabular}{lll}
\textbf{Number} & \textbf{Entity} & \textbf{Symbol} \\ \hline
1               & Axis Bank       & AXISBANK.BO     \\
2               & Bank of Baroda  & BANKBARODA.BO   \\
3               & Federal Bank    & FEDERALBNK.BO   \\
4               & HDFC Bank       & HDFCBANK.BO     \\
5               & ICICI Bank      & ICICIBANK.BO    \\
6               & Indus Ind Bank  & INDUSINDBK.BO   \\
7               & Kotak Mahindra  & KOTAKBANK.BO    \\
8               & PNB             & PNB.BO          \\
9               & SBI             & SBIN.BO         \\
10              & Yes Bank        & YESBANK.BO     \\ \hline
\end{tabular}
\end{table}

\begin{figure}[htp]
\centering

  \includegraphics[width=0.7\textwidth]{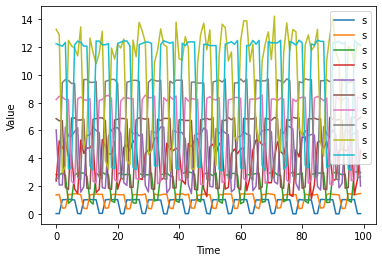}%

\caption{Time series in the Activities dataset without normalization, first 100 samples.}
\label{fig:activities}
\end{figure}
\subsection{The Activities dataset}\label{s:activities}
The Activities dataset is a synthetic dataset created resembling weekly activities with five days of high activity and two days of low activity. The dataset has ten time series with $3\,584$ samples per series. Initially, a pattern of five ones followed by two zeros was repeated to obtain a length of $3\,584$ samples. The series was added a slope of $0.0001$. The original series was circularly rotated for the remaining series in the dataset, to which noise was added, and each sequence was arbitrarily scaled, so that the peak-to-peak amplitude of each series was different, see Figure \ref{fig:activities}.

\subsection{Datasets preparation and partition}\label{s:partition}
Following section \ref{s:datap}, every time series was normalized between 0 and 1. We used a window of size $w=60$ days. We tested for $f=1$ and $f=20$ steps ahead. We used the last 251 samples of each time series for testing. We selected arbitrarily the first time series of each dataset for training our LSTM and GRU networks.

\subsection{Results}\label{s:results}
The results are presented in Tables \ref{tabA1},  \ref{tabA20},  \ref{tabB1}, and  \ref{tabB20}. Close-to-zero RMSE and close-to-one DA are preferred. On the Activities dataset, two-tailed Mann-Whitney tests show that for 1-step ahead forecasts, RMSE achieved by any RNN is significantly lower than that delivered by the baseline (LSTM  \& Baseline: $U = 19, n = 10, 10, P<0.05$. GRU  \& Baseline: $U = 0, n = 10, 10, P<0.05$). In addition, GRU delivers significantly lower RMSE than LSTM ($U = 91, n = 10, 10, P<0.05$). In terms of DA, both RNN perform equally well and significantly outperform the baseline. For 20-step ahead forecasts, again both RNNs achieve significantly lower RMSE than the baseline (LSTM  \& Baseline: $U = 0, n = 10, 10, P<0.05$. GRU  \& Baseline: $U = 0, n = 10, 10, P<0.05$). This time, LSTM achieves lower RMSE than GRU ($U = 10, n = 10, 10, P<0.05$) and higher DA ($U = 81, n = 10, 10, P<0.05$).

\begin{figure}[htp]
\centering
\subfloat[Activities dataset. \label{fig:subim1}]{%
  \includegraphics[width=0.5\textwidth]{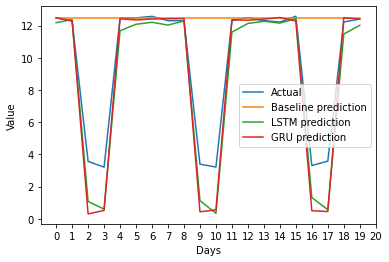}%
}\hfil
\subfloat[BANKEX dataset.\label{fig:subim2}]{%
  \includegraphics[width=0.5\textwidth]{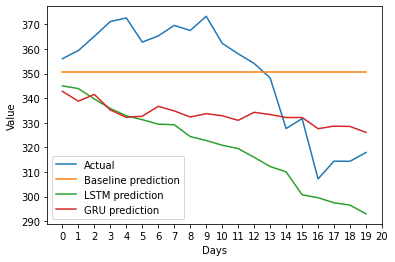}%
}

\caption{Examples of 20-step ahead forecast.}
\label{fig:20}

\end{figure}

\begin{figure}
   \centering
   \includegraphics[width=1\linewidth]{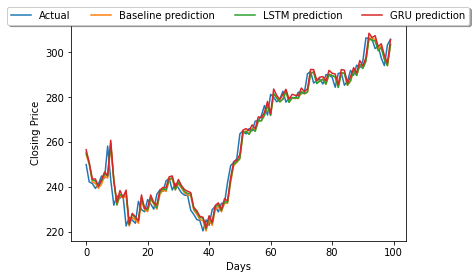}
   \caption{Example of 1-step ahead forecast. Actual and predicted closing price over the first 100 days of the test set Yes Bank. Closing Price in Indian Rupee (INR).}
   \label{fig:pred_example} 
\end{figure}

On the BANKEX dataset, two-tailed Mann-Whitney tests show that for 1-step ahead forecasts there is no difference among approaches considering RMSE (LSTM  \& Baseline: $U = 51, n = 10, 10, P>0.05$. GRU  \& Baseline: $U = 55, n = 10, 10, P>0.05$. LSTM  \& GRU: $U = 49, n = 10, 10, P>0.05$). Similar results are found for 20-step ahead forecasts (LSTM  \& Baseline: $U = 76, n = 10, 10, P>0.05$. GRU  \& Baseline: $U = 67, n = 10, 10, P>0.05$. LSTM  \& GRU: $U = 66, n = 10, 10, P>0.05$). DA results are consistent with those obtained for RMSE. Figure \ref{fig:pred_example} presents an example of 1-step ahead forecasts and Figure \ref{fig:20} shows examples of 20-step ahead forecasts. Visual inspection helps understand the results.

\begin{table}[]
\centering
\caption{One-step ahead forecast on Activities dataset. }\label{tabA1}
%\begin{tabular}{lllllll}
\begin{tabular}{c|ccc|ccc} \hline
              & \multicolumn{3}{c}{\textbf{RMSE}}                & \multicolumn{3}{|c}{\textbf{DA}}                  \\ \hline
              & \textbf{LSTM} & \textbf{GRU} & \textbf{Baseline} & \textbf{LSTM} & \textbf{GRU} & \textbf{Baseline} \\  \hline
\textbf{Mean} &     0.2949          &     0.1268         &       0.3730            &      0.6360         &    0.6236          &     0.4212              \\
\textbf{SD}   &      0.0941         &     0.0425 & 0.0534 & 0.0455 & 0.0377 & 0.0403         \\ \hline   
\end{tabular}
\end{table}

\begin{table}[]
\centering
\caption{Twenty-step ahead forecast on Activities dataset.}\label{tabA20}
%\begin{tabular}{lllllll}
\begin{tabular}{c|ccc|ccc} \hline
              & \multicolumn{3}{c}{\textbf{RMSE}}                & \multicolumn{3}{|c}{\textbf{DA}}                  \\ \hline
              & \textbf{LSTM} & \textbf{GRU} & \textbf{Baseline} & \textbf{LSTM} & \textbf{GRU} & \textbf{Baseline} \\  \hline
\textbf{Mean} &     0.1267 & 0.2048& 0.4551& 0.6419& 0.6261&
       0.4805              \\
\textbf{SD}   &      0.0435& 0.0683& 0.0678& 0.0331& 0.0255&       0.0413        \\ \hline   
\end{tabular}
\end{table}
\begin{table}[]
\centering
\caption{One-step ahead forecast on BANKEX dataset.}\label{tabB1}
%\begin{tabular}{lllllll}
\begin{tabular}{c|ccc|ccc} \hline
              & \multicolumn{3}{c}{\textbf{RMSE}}                & \multicolumn{3}{|c}{\textbf{DA}}                  \\ \hline
              & \textbf{LSTM} & \textbf{GRU} & \textbf{Baseline} & \textbf{LSTM} & \textbf{GRU} & \textbf{Baseline} \\  \hline
\textbf{Mean} &     0.0163 & 0.0163& 0.0161& 0.4884   & 0.4860  &  0.4880    \\
\textbf{SD}   &      0.0052& 0.0056& 0.0056& 0.0398& 0.0385& 0.0432     \\ \hline   
\end{tabular}
\end{table}

\begin{table}[]
\centering
\caption{Twenty-step ahead forecast on BANKEX dataset.}\label{tabB20}
%\begin{tabular}{lllllll}
\begin{tabular}{c|ccc|ccc} \hline
              & \multicolumn{3}{c}{\textbf{RMSE}}                & \multicolumn{3}{|c}{\textbf{DA}}                  \\ \hline
              & \textbf{LSTM} & \textbf{GRU} & \textbf{Baseline} & \textbf{LSTM} & \textbf{GRU} & \textbf{Baseline} \\  \hline
\textbf{Mean} &    0.0543& 0.0501&0.0427&0.5004& 0.5004& 0.4969    \\
\textbf{SD}   &   0.0093& 0.0064&0.0113& 0.0071&0.0087& 0.0076    \\ \hline   
\end{tabular}
\end{table}
       
%\newpage
\section{Discussion}\label{s:discussion}
The motivation for developing a reproducible and open-source framework for time series forecasting relates to our experience in trying to reproduce previous work \cite{siami2018comparison,balaji2018applicability}. We found it challenging to find implementations that are simple to understand and replicate. In addition, datasets are not available. We discontinued comparisons with \cite{balaji2018applicability}, since the dataset we collected was slightly different, and we were unsure if the reported results referred to normalized values or not. If the algorithms are described but the implementations are not available, a dataset is necessary to compare forecasting performance between two algorithms, such that a statistical test can help determine if one algorithm is significantly more accurate than the other \cite[p. 580-581]{alpaydin2014}. 

\section{Conclusion}\label{s:conclusions}
We proposed a method for time series forecasting based on LSTM and GRU and showed that these networks can be successfully trained with a single time series to deliver forecasts for unseen time series in a dataset containing patterns that repeat, even with certain variation. Once a network is properly trained, it can be used to forecast other series in the dataset if adequately prepared. We tried and varied several hyperparameters. On sequences, such as those resembling weekly activities that repeat with certain variation, we found an appropriate setting. However, we failed to find an architecture that would outperform a baseline on stock market data. Therefore, we assume that we either failed at optimizing the hyperparameters of the networks, or we would need extra information that is not reflected in stock market series alone. For future work, we plan to benchmark different forecasting methods against the method presented here. In particular, we want to evaluate statistical methods as well as other machine learning methods that have demonstrated strong performance on forecasting tasks \cite{makridakis2022m5}. We release our code as well as the dataset used in this study to allow this research to be reproducible. 

\subsubsection{Acknowledgments}
We would like to thank the anonymous reviewers for their valuable observations. 

%\bibliographystyle{splncsnat}

\bibliographystyle{splncs04}
\bibliography{References.bib}
\end{document}